\documentclass[10pt,twocolumn,letterpaper]{article}

\usepackage{cvpr}
\usepackage{times}
\usepackage{epsfig}
\usepackage{graphicx}
\usepackage{amsmath}
\usepackage{amssymb}
\usepackage{multirow}
\usepackage{tabularx}
\usepackage{algorithm}
\usepackage{algorithmic}

\def\E{{\rm E}}
\def\N{{\rm N}}

\def\G{{\cal G}}

\def\hY{\hat{Y}}
\def\hZ{\hat{Z}}

\usepackage[pagebackref=true,breaklinks=true,letterpaper=true,colorlinks,bookmarks=false]{hyperref}

\cvprfinalcopy 

\def\cvprPaperID{9926} 

\ifcvprfinal\fi
\begin{document}

\title{Inducing Hierarchical Compositional Model by Sparsifying Generator Network}

\author{Xianglei Xing$^{1}$, Tianfu Wu$^{2}$, Song-Chun Zhu$^{3}$, Ying Nian Wu$^{3}$\\
$^1$College of Automation, Harbin Engineering University, Harbin 150001, China\\
$^2$Department of Electrical and Computer Engineering, NC State University, North Carolina 27695\\
$^3$Department of Statistics, University of California, Los Angeles,  California 90095\\
{\tt\small xingxl@hrbeu.edu.cn, tianfu\_wu@ncsu.edu,\{sczhu,ywu\}@stat.ucla.edu}
}

\maketitle

\begin{abstract}
 This paper proposes to learn hierarchical compositional AND-OR model for interpretable image synthesis by sparsifying the generator network. The proposed method adopts the scene-objects-parts-subparts-primitives hierarchy in image representation. A scene has different types (i.e., OR) each of which consists of a number of objects (i.e., AND). This can be recursively formulated across the scene-objects-parts-subparts hierarchy and is terminated at the primitive level (e.g., wavelets-like basis).
To realize this AND-OR hierarchy in image synthesis, we learn a generator network that consists of the following two components: (i) Each layer of the hierarchy is represented by an over-complete set of convolutional basis functions.  Off-the-shelf convolutional neural architectures are exploited to implement the hierarchy. (ii) Sparsity-inducing constraints are introduced in end-to-end training, which induces a sparsely activated and sparsely connected AND-OR model  from the initially densely connected generator network. A straightforward sparsity-inducing constraint is utilized, that is to only allow the top-$k$ basis functions to be activated at each layer (where $k$ is a hyper-parameter). The learned basis functions are also capable of image reconstruction to explain the input images.
In experiments, the proposed method is tested on four benchmark datasets. The results show that meaningful and interpretable hierarchical representations are learned with better qualities of image synthesis and reconstruction obtained than baselines.
\end{abstract}

\section{Introduction}
Remarkable recent progress on image synthesis~\cite{goodfellow2014generative,brock2018large,karras2018style,coopnets,arjovsky2017wasserstein,xing2019unsupervised} has been  made  using deep neural networks (DNNs)~\cite{LeCunCNN,AlexNet}. Most efforts focus on developing sophisticated architectures and training paradigms for sharp and photo-realistic image synthesis~\cite{lucic2017gans,BigGAN,karras2018style}. Although high-fidelity images can be generated, the internal synthesizing process via DNNs is still largely viewed as a black-box, thus potentially hindering the long-term applicability in eXplainable AI (XAI)~\cite{XAI}. More recently, the generative adversarial network (GAN) dissection method~\cite{bau2019gandissect} has been proposed to identify  internal neurons  in pre-trained GANs that show interpretable meanings  using a separate annotated dataset in a post-hoc fashion.

\begin{figure*}[t!]
	\begin{center}
		\includegraphics[width=2.0\columnwidth]{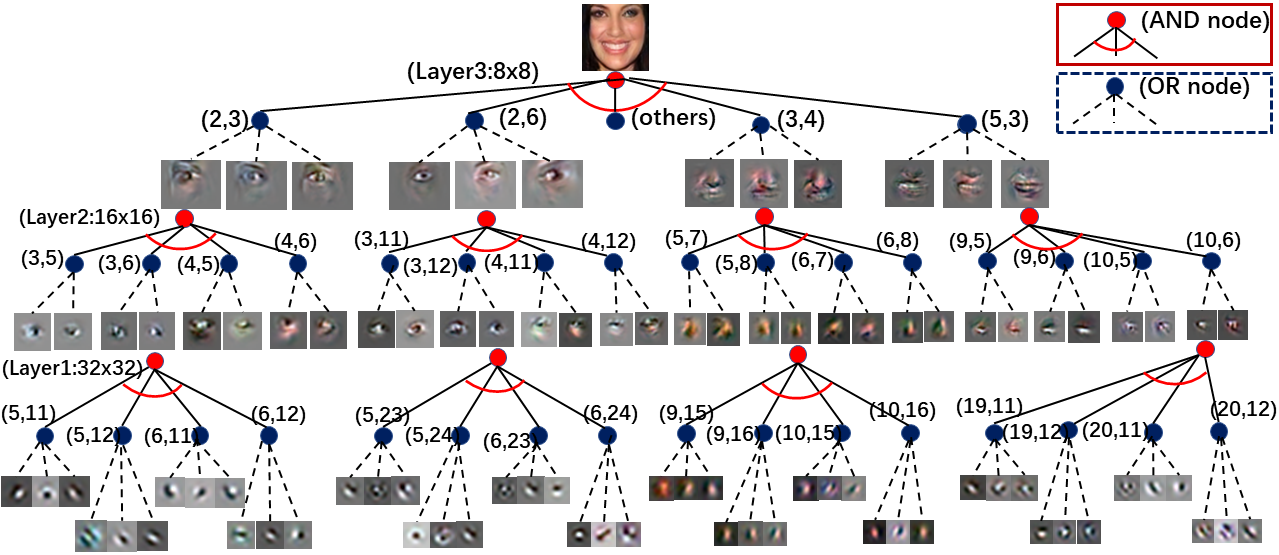}\\
		\caption{An example of AND-OR tree learned for face synthesis at a resolution of $64\times 64$. For clarity, we only show 3 layers (out of the total 5 layers). 
			The AND-OR tree is illustrated from Layer 3 (the part/composite part level) to Layer 1 (the primitive level). The spatial resolution of the feature map at Layer 3 is $8\times 8$. The entire grid is interpreted by a AND-node. Each position $(x, y)$ in the $8\times 8$ grid is interpreted by an OR-node such as the eye node at the positions $(2,3)$ and $(2, 6)$. Similarly, Layer 2 and Layer 1 can be interpreted w.r.t. the AND-OR compositions. The activated basis functions have semantically meaningful interpretations at Layer 3 and Layer 2. Layer 1 shows the learned primitives covering the classic Gabor-like wavelets and the blob-like primitives. See text for detail. Best viewed in color.}
		\label{fig:aot_face} \vspace{-5mm}
	\end{center}
\end{figure*}

In this paper, we focus on learning  interpretable models for unconditional image synthesis from scratch with explicit hierarchical representations. Interpretable image synthesis means that the internal image generation process can be explicitly unfolded through meaningful basis functions at different layers which are learned via end-to-end training  and which conceptually reflect the hierarchy of scene-objects-parts-subparts-primitives.  A scene has different types (i.e., OR) each of which consists of a number of objects (i.e., AND). This can be recursively formulated across the scene-objects-parts-subparts hierarchy and is terminated at the primitive level (e.g., wavelets-like basis functions).  Figure~\ref{fig:aot_face} shows an example of the AND-OR tree learned from scratch for explaining the generation of a face image.

The hierarchy of scene-objects-parts-subparts-primitives is at the stem of image grammar models~\cite{Geman_CompositionSystems,zhu2007stochastic}. The AND-OR compositionality has been applied in image and vision tasks~\cite{zhu2007stochastic}. With the recent resurgence of  DNNs~\cite{LeCunCNN,AlexNet} and the more recent DNN-based image synthesis frameworks such as the widely used Generative Adversarial Networks (GANs)~\cite{goodfellow2014generative} and Variational Auto-Encoder (VAE) methods \cite{kingma2013auto,higgins2016beta}, the hierarchy is usually assumed to be modeled implicitly in DNNs. Due to dense connections between consecutive layers in traditional DNNs, they often learn dense compositional patterns of how entities in a layer are formed from ``smaller" ones in the layer right below it.

The sparsity principle has played a fundamental role in high-dimensional statistics, machine learning, signal processing and AI. In particular, the sparse coding scheme~\cite{olshausen1996emergence} is an important principle for understanding the visual cortex. By imposing sparsity constraints on the coefficients of a linear generative model, \cite{olshausen1997sparse} learned Gabor-like wavelets that resemble the neurons in the primary visual cortex (V1). Since then, there have been many important developments on sparse coding presented in the literature before the resurgence of DNNs.
With the remarkable successes of sparse coding models, it is not unreasonable to assume that a top-down generative model of natural images should be based on the linear sparse coding model, or incorporate the sparse coding principle at all of its layers.
However, developing a top-down sparse coding model that can generate, rather than merely reconstruct,  photo-realistic natural image patterns has proven to be a difficult task~\cite{ishwaran2005spike}, mainly due to the difficulty of selecting and fitting sparse basis functions to each image.
	
In this paper, we take a step forward by rethinking dense connections between consecutive layers in the generator network. We propose to ``re-wire" them sparsely for explicit modeling of the hierarchy of scene-objects-parts-subparts-primitives in image synthesis (see Figure~\ref{fig:aot_face}).  To realize the ``re-wiring", we integrate the sparsity principle into the generator network in a simple yet effective and adaptive way:  (i) Each layer of the hierarchy is represented by an over-complete set of basis functions. The basis functions are instantiated using convolution in order to be translation covariant. Off-the-shelf convolutional neural architectures are then exploited to implement the hierarchy such as generator networks used in GANs. (ii) Sparsity-inducing constraints are introduced in end-to-end training which facilitates a sparsely connected AND-OR model to emerge from initially densely connected convolutional neural networks. A straightforward sparsity-inducing constraint is utilized, that is to only allow the top-$k$ basis functions to fire at each layer (where $k$ is a hyper-parameter).  By doing so, we can harness the highly expressive modeling capability and the end-to-end learning flexibility of generator network, and the interpertability  of the explicit compositional hierarchy.

\section{Related work}
Sparse-autoencoders~\cite{ng2011sparse,makhzani2013k,hosseini2015deep} were proposed for effective feature representations and these representations can improve the performance of the classification task. The sparsity constrains are designed and encouraged by the Kullback-Leibler divergence between the Bernoulli random variables \cite{ng2011sparse}, $l_1$ penalty on the normalized features \cite{ngiam2011sparse}, and winner-takes-all principle \cite{makhzani2015winner}. However, these methods do not have the ability to generate new data. Lee \cite{lee2009convolutional,huang2012learning} proposed a convolutional deep belief network which employs sparsity regularization and probabilistic max-pooling to learn hierarchical representations. However, the learning is difficult and computationally expensive for training the deep belief nets. Zeiler \cite{zeiler2010deconvolutional,zeiler2011adaptive} proposed the deconvolutional networks to learn the low and mid-level image representations based on the convolutional decomposition of images under a sparsity constrain.
However, for the aforementioned methods, the hierarchical representations have to be learned layer by layer, that is to first train the bottom layer of the network and then fix the learned layer and train the upper layers one by one. Moreover, the above methods usually work on  gray-level images or the gradient images which are preprocessed by removing the low frequency information and highlighting the structure information. Unlike the above methods, the proposed method can directly work on the raw color images without any preprocessing. The proposed model can simultaneously learn meaningful hierarchical representations, generate realistic images and reconstruct the original images.

\noindent\textbf{Our contributions.} This paper makes three main contributions to the field of generative learning: (i) It proposes interpretable image synthesis that unfolds the internal generation process via a hierarchical AND-OR model of semantically meaningful nodes. (ii) It presents a simple yet effective sparsity-inducing method that facilitates a hierarchical AND-OR model of sparsely connected nodes to emerge from an initial network of dense connections between consecutive layers. (iii) It shows that meaningful hierarchical representations can be learned in an end-to-end manner in image synthesis with better qualities than baselines.

\section{Sparsifying generator network}

\subsection{Image synthesis and model interpretability}
From the viewpoint of top-down generative learning in image synthesis, we start with a $d$-dimensional latent code vector $Z=(z_i, i=1,\cdots, d)$ consisting of $d$ latent factors. We usually assume $Z\sim \mathcal{N}(0, I_d)$, where $I_d$ denotes the $d$-dimensional identity matrix. In GANs and VAE, generator networks are used to implement the highly non-linear mapping from a latent code vector $Z$ to a synthesized image, denoted by $Y$ which lies in a $D$-dimensional image space (i.e., $D$ equals the product of the spatial dimensions, width and height of an image, and the number of chromatic channels such as $3$ for RGB images). The generator network is thus seen as non-linear extension of factor analysis~\cite{han2017alternating}. The model has the following form:
\begin{align}
    &Y = g(Z; \Theta) + \epsilon, \label{eq:factor}\\
\nonumber    &Z\sim \mathcal{N}(0, I_d),\quad  \epsilon\sim \mathcal{N}(0, \sigma^2 I_D), \quad d < D,
\end{align}
where $\epsilon$ is the observational errors assumed to be Gaussian white noises, $g(\cdot)$ represents the  generator network and $\Theta$ collects  parameters from all layers.

\begin{figure}[t!]
	\begin{center}
		\includegraphics[width=1.0\columnwidth]{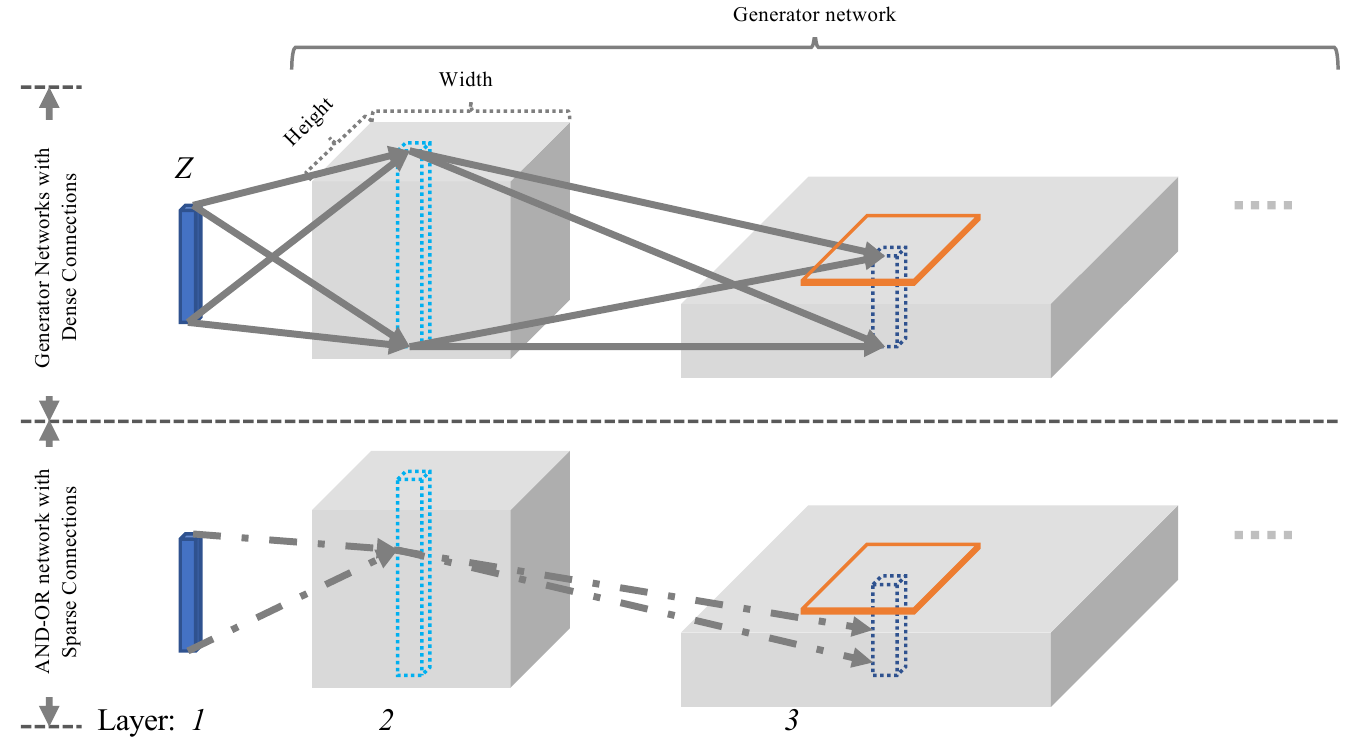}\\
		\caption{\textit{Top}: traditional generator networks with dense connections (solid arrows) between consecutive layers, which are widely used in GANs and VAE.  \textit{Bottom}: the proposed AND-OR models with sparse connections (dashed arrows). See text for detail.}
		\label{fig:network} \vspace{-5mm}
	\end{center}
\end{figure}

As illustrated at the top of Figure~\ref{fig:network}, dense connections between consecutive layers are learned in the vanilla generator network, which we think is the main drawback that hinders explicit model intepretability. We explore and exploit the AND-OR compositionality in image synthesis by learning to rewire the connections sparsely and to unfold the internal image generation process in an interpretable way, as illustrated at the bottom of Figure~\ref{fig:network}.

\subsection{The induced AND-OR model}
Without loss of generality, consider a simple hierarchy of object(O)-part(P)-primitive/basis(B) that generates RGB images.  Start with the latent code vector $Z\in R^d$,  we have,
\begin{align}
    \text{Hierarchy: } &Z\rightarrow &&O\rightarrow &&P\rightarrow &&B\rightarrow &&Y, \\
    \text{Layer Index: } &1, &&2, &&3, &&4, &&5.
\end{align}
For example, Figure~\ref{fig:network} illustrates the computing flow from Layer 1 to Layer 3.

The symbol $O$ in the hierarchy is grounded in an internal $(H_2\times W_2\times d_2)$-dimensional space, which can be treated as $H_2\times W_2$ $d_2$-dimensional vectors when instantiated. Similarly, the symbols $P$ and $B$ will be instantiated as $H_3\times W_3$ $d_3$-dimensional vectors and $H_4\times W_4$ $d_4$-dimensional vectors respectively. $Y$ is a generated RGB image of size $H_5\times W_5\times 3$.

To better show how to facilitate the sparse connections to emerge from the dense ones, we look at the computing flow using the lens of vector-matrix multiplication~\cite{VectorMatrix19}. In the vanilla generator network, consider a $d_l$-dimensional vector, $U$ in Layer $l$, it connects to a set of $d_{l+1}$-dimensional vectors, $V_i$'s in Layer $l+1$. Let $ch(U)$ be the set of indices of the vectors in Layer $l+1$ which connect with $U$ (i.e., $U$'s child nodes).  We have
\begin{equation}
    V_i(U) = W_{d_{l+1}\times d_l} \cdot U + b,\quad i\in ch(U),
\end{equation}
where $V_i(U)$ means the contribution of $U$ to $V_i$ since there may be other vectors $U'$ in Layer $l$ connecting to $V_i$ too. $W_{d_{l+1}\times d_l}$ is the transformation matrix and $b$ the bias vector. Consider Layer 1 to Layer 2 ($Z\rightarrow O$), $U=Z$ is connected with all vectors $V_i$'s in $O$ with different $W_{d_{l+1}\times d_l}$'s and $b$'s. Consider Layer 2 to Layer 3 ($O\rightarrow P$), convolution is usually used, so each $U$ only connects to vectors $V_i$'s locally, and $W_{d_{l+1}\times d_l}$'s and $b$'s are shared among different $U$'s.

Denote by $pr(V_i)$ the set of indices of vectors in Layer $l$ connecting with $V_i$. In the vanilla generator network,
\begin{equation}
    V_i = act(\sum_{j\in pr(V_i)} V_i(U_j)),
\end{equation}
where $act(\cdot)$ stands for activation function such as the ReLU function~\cite{AlexNet}.

In the proposed method, we compute $V_i$ by,
\begin{equation}
   V_i = S(act(\sum_{j\in pr(V_i)} V_i(U_j)); k_{l+1}),
\end{equation}
where $S(\cdot; k_{l+1})$ is the sparsity-inducing function.  From symbol $Z$ to $O$, we apply the sparsity-inducing function along the $d_{l+1}$ dimension and  retain the top $k_{l+1}$ out of $d_{l+1}$ elements in the resulting vector in terms of the element values. In the subsequent layers, we apply it along the spatial domain across  the $d_{l+1}$ dimensions individually. By doing so, the resulting vectors at different locations will have different sparsity ratios. The  $k_{l}$'s are  hyper-parameters. We usually set $d_l > d_{l+1}$ and $k_l < k_{l+1}$, that is, Layer $l$ has higher sparsity degree than lower Layer $l+1$.

With  sparsity-inducing functions,  image synthesis  is fundamentally changed in terms of representation. The internal generation process is also much easier to unfold. The AND-OR model then emerges from the vanilla dense connected generator network. We can rewrite Eqn.~\ref{eq:factor} as
\begin{align}
    &Y = g(Z; \Theta, \mathbf{k}) + \epsilon, \label{eq:factor_sparse}\\
\nonumber    &Z\sim \mathcal{N}(0, I_d),\quad  \epsilon\sim \mathcal{N}(0, \sigma^2 I_D), \quad d < D,
\end{align}
where the sparsity hyper-parameters $\mathbf{k} = \{k_l; l=1, \cdots, L\}$.
We summarize the proposed AND-OR model for image synthesis as follows.

\textbf{Layer 1 to Layer 2: $Z\rightarrow O$}. The latent code vector $Z$ is represented by a root OR-node (non-terminal symbol),
\begin{equation}
    Z \xrightarrow[]{\text{OR}} z_1 | z_2 | \cdots | z_i | \cdots, \quad z_i \stackrel{\text{i.i.d.}}{\sim} \mathcal{N}(0, I_d),
\end{equation}
where  $a | b$ denotes OR switching between symbols $a$ and $b$ (i.e., instantiated latent code vectors that generate different object images).

Each instantiated latent code vector $z_i$ is then mapped to an object instance AND-node $O_i$.
The object instance AND-node $O_i$ represents the object-part decomposition in the lattice $\Lambda_2$ (of size $H_2\times W_2$). We have,
\begin{equation}
    O_i \xrightarrow[]{\text{AND}} o_{i,1} \cdot o_{i,2} \cdot \cdots \cdot o_{i,j} \cdot \cdots \cdot o_{i,N_P},
\end{equation}
where $a\cdot b$ represents the composition between symbols $a$ and $b$. $N_P$ is the number of part symbols. The object-part decomposition is usually done in the spatial domain. For example, if the support domain for $o_i$'s is $4\times 4$, we will have at most $16$ parts. We could use $4$ parts if we further divide the $4\times 4$ domain into $2\times 2$ blocks.

Each $o_{i,j}$ is then represented by an OR-node in the $d_2$-dimensional vector space indicating the sparse selection  among $\binom{d_2}{k_2}$ candidates. When instantiated, we have part AND-node $o_{i,j}(k_2)$.

\textbf{Layer 2 to Layer 3: $O\rightarrow P$}.
Each part AND-node $o_{i,j}(k_2)$ is decomposed into a number of $M$ child part type OR-nodes,
\begin{equation}
    o_{i,j}(k_3) \xrightarrow[]{\text{AND}} P_{i,j,1} \cdot P_{i,j,2} \cdot \cdots \cdot P_{i,j,M},
\end{equation}
where $M$ is determined by the kernel size when convolution is used to compute Layer 3 from Layer 2.

Similarly, each part type OR-node $P_{i,j,t}$ is grounded in the $d_3$-dimensional vector space indicating the sparse selection  among $\binom{d_3}{k_3}$ candidates. When instantiated, we have part-primitive AND-node. Then, the AND-OR is recursively formulated in the downstream layers. Now, let us look at Figure~\ref{fig:network} again, for each instantiated $z_i$, we can follow the sparse connections and visualize the encountered kernel symbols (see Figure~\ref{fig:aot_face}).

\subsection{Learning and inference}
The proposed AND-OR model can still utilize off-the-shelf end-to-end learning framework since the sparsity-inducing functions do not change the formulation (Eqn.~\ref{eq:factor_sparse}). We adopt the alternating back-propagation learning framework proposed in~\cite{han2017alternating}.

Denote by  $\{Y_i,i = 1,\dots,N\}$ the training dataset consisting of $N$ images (e.g., face images). The learning objective is to maximize the observed data log-likelihood
	\begin{eqnarray}
		L(\Theta) &=& \frac{1}{N}\sum_{i=1}^N\log P(Y_i; \Theta, \mathbf{k})  \nonumber \\
		&=& \frac{1}{N} \sum_{i=1}^N \log \int P(Y_i, Z_i; \Theta, \mathbf{k})dZ_i,\label{eq:llh}
	\end{eqnarray}
	where the latent vector $Z_i$ for an observed data $Y_i$ is integrated out, and $P(Y_i, Z_i; \Theta, \mathbf{k})$ is the complete-data likelihood. The gradient of $L(\Theta)$ is computed as follows
	\begin{eqnarray}
		&&  \frac{\partial}{\partial \Theta} \log P(Y; \Theta, \mathbf{k})  \nonumber \\
		&=&  \frac{1}{P(Y; \Theta, \mathbf{k})}  \frac{\partial}{\partial \Theta}  \int P(Y, Z; \Theta, \mathbf{k}) dZ \nonumber \\
		&=& \E_{P(Z|Y; \Theta, \mathbf{k})} \left[ \frac{\partial}{\partial \Theta} \log P(Y, Z;  \Theta, \mathbf{k})\right].  \label{eq:EM}
	\end{eqnarray}
	In general, the expectation in Eqn.\ref{eq:EM} is analytically intractable. Monte Carlo average is usually adopted in practice with samples drawn from the posterior $P(Z|Y; \Theta, \mathbf{k})$ by the Langevin dynamics,
	\begin{eqnarray}
		Z_{\tau + 1} = Z_\tau + \frac{\delta^2}{2}  \frac{\partial}{\partial Z}  \log P(Z_\tau, Y;  \Theta, \mathbf{k}) + \delta {\cal E}_\tau,  \label{eq:LangevinG}
	\end{eqnarray}
	where  $\tau$ indexes the time step, $\delta$ is the step size, and ${\cal E}_\tau$ denotes the noise term, ${\cal E}_\tau \sim \N(0, I_d)$.
	
	Based on Eqn.~\ref{eq:factor_sparse}, the complete-data log-likelihood is computed by
	\begin{eqnarray}
		\log p(Y, Z; \Theta, \mathbf{k}) = \log \left[p(Z) p(Y|Z; \Theta, \mathbf{k}) \right]\nonumber\\
		= -\frac{1}{2\sigma^2} \|Y - g(Z; \Theta, \mathbf{k})\|^2 -\frac{1}{2} \|Z\|^2 + C
	\end{eqnarray}
	where $C$ is a constant term independent of $Z$ and $Y$. It can be shown that, given sufficient  transition steps, the $Z$  obtained from this procedure follows the posterior distribution. For each training example $Y_i$, we run the Langevin dynamics in Eqn.\ref{eq:LangevinG} to get the corresponding posterior sample $Z_i$.  The sample is then used for gradient computation in Eqn.\ref{eq:EM}. The parameters $\Theta$ are then learned through Monte Carlo approximation,
	\begin{eqnarray}
		\frac{\partial}{\partial \theta}L(\Theta) \approx \frac{1}{N}\sum_{i=1}^N \frac{\partial}{\partial \Theta} \log p(Y_i, Z_i; \Theta, \mathbf{k})\nonumber\\
		=\frac{1}{N}\sum_{i=1}^N \frac{1}{\sigma^2} (Y_i-g(Z_i; \Theta, \mathbf{k}))\frac{\partial}{\partial \Theta}g(Z_i; \Theta, \mathbf{k}).
		\label{eq:weight}
	\end{eqnarray}

\subsection{Energy-based model as a critic}
It is well known that using squared Euclidean distance alone to train  generator networks often yields blurry reconstruction results, since the precise location information of details may not be preserved, and the $L_2$ loss in the image space leads to averaging effects among all likely locations.  In order to improve the quality, we recruit an energy-based model as a critic of the generator model which serves as an actor. The energy-based model is in the form of exponential tilting of a reference distribution
	\begin{eqnarray}
		P(Y; \Phi) = \frac{1}{Z(\Phi)} \exp\left[- f(Y; \Phi)\right] q(Y),
	\end{eqnarray}
	where $f(Y; \Phi)$ is parameterized by a bottom-up ConvNet which maps an image $Y$ to the feature statistics or energy, $
	Z(\Phi) = \int  \exp\left[ f(Y; \Phi)\right] q(Y) dY
	= \E_{q}\{ \exp[f(Y; \Phi)]\}
	$ is the normalizing constant, and $q(Y)$ is the reference distribution such as Gaussian white noise,
	\begin{eqnarray}
		q(Y) = \frac{1}{(2\pi \sigma^2)^{D/2}} \exp \left[ - \frac{\|Y\|^2}{2 \sigma^2}\right].
	\end{eqnarray}

 Let $P(Y)$ be the underlying data distribution. Expectation with respect to $P(Y)$ is taken to be the average over training examples. We jointly learn the generator model and the energy-based model by introducing the following cross-entropy triangle, under a unified probabilistic framework,
 \begin{align}
  \nonumber & \min_\Theta \max_\Phi T(\Theta, \Phi), \\
    & T(\Theta, \Phi) = H(P(Y), P(Y;\Theta,\mathbf{k}))    \label{eq:3kl}\\
\nonumber	&- H(P(Y), P(Y;\Phi)) + H(P(Y;\Theta, \mathbf{k}),P(Y;\Phi)),
\end{align}
where $H(P,Q) = -\E_P \left[ \log Q \right]$ denotes the cross-entropy between the two distributions.  $H(P(Y), P(Y;\Theta,\mathbf{k}))$ and
$H(P(Y), P(Y;\Phi))$ lead to the maximum likelihood learning of the two models respectively, while $H(\Theta, \Phi) = H(P(Y;\Theta, \mathbf{k}),P(Y;\Phi))$ connects and modifies the learning of the two models.  $H(P(Y;\Theta, \mathbf{k}),P(Y;\Phi))$ causes the following effect: the energy-based model criticizes the generator model by assigning lower energies to the observed examples than the synthesized examples. The generator model then improves its synthesis by lowering the energies of the synthesized examples.

Specifically,  to update $\Theta$, minimizing the first term is equivalent to maximizing Eqn.~\ref{eq:llh}. The third term in Eqn.\ref{eq:3kl} can be written as
\begin{align}
    \nonumber  H(&P(Y;\Theta, \mathbf{k}), P(Y;\Phi)) = -\E_{p(Y;\Theta,\mathbf{k})} \log P(Y;\Phi) \\
    &= -\E_{Z\sim p(Z)} \log P(g(Z;\Theta,\mathbf{k});\Phi).
\end{align}
\begin{eqnarray}
		&&-\frac{\partial}{\partial \Theta} \E_{Z\sim p(Z)} \log P(g(Z;\Theta,\mathbf{k}); \Phi) \nonumber \\
        &&\approx   \frac{1}{N} \sum_{i=1}^{N} \frac{\partial}{\partial \Theta} f(g(Z_i; \Theta,\mathbf{k}); \Phi),
		\label{eq:crossen}
\end{eqnarray}
which seeks to make the synthesized $(g(Z_i; \Theta,\mathbf{k}))$ to have low energies.

To update $\Phi$, we have
   \begin{align}
	\nonumber	-\frac{\partial}{\partial \Phi}T(\Theta, \Phi) \approx & \frac{1}{N} \sum_{i=1}^{N} \frac{\partial}{\partial \Phi} f(Y_i; \Phi) - \\   & \frac{1}{N} \sum_{i=1}^{N} \frac{\partial}{\partial \Phi} f(g(Z_i; \Theta,\mathbf{k}); \Phi), \label{eq:lD2}
	\end{align}
which seeks to make the energies of the observed $(Y_i)$ to be lower than the energies of the synthesized $(g(Z_i; \Theta,\mathbf{k}))$.

 Algorithm~\ref{alg:learning} presents the detail of learning and inference.

 \begin{algorithm}[t]
 	\caption{Learning and Inference Algorithm}
 	\label{alg:learning}
 	\begin{algorithmic}[1]
 		
 		\REQUIRE ~~\\
 		(1) training examples $\{Y_i, i=1,...,N\}$ \\
 		(2) network architectures and sparsity-inducing hyper-parameters $\mathbf{k}$ (see Table~\ref{table:network})\\
 		(3) Langevin steps $l_\G$ and learning iterations $T$
 		
 		\ENSURE~~\\
 		(1) estimated parameters $\Theta$ and $\Phi$ \\
 		(2) synthesized examples $\{\hY_i,  i= 1, ..., {N}\}$ \\
 		
 		\STATE Let $t\leftarrow 0$,  initialize $\Theta$ and $\Phi$.
 		\REPEAT
 		\STATE {\bf Step 1}: For $i = 1, ..., N$, generate $\hZ_i \sim \N(0, I_d)$, and generate $\hY_i = g(\hZ_i; \Theta^{(t)},\mathbf{k}) $. Update $\Phi^{(t+1)} = \Phi^{(t)} + \gamma_t \frac{\partial}{\partial \Phi}T(\Theta^{(t)}, \Phi^{(t)})$,  where $\frac{\partial}{\partial \Phi}T(\Theta^{(t)}, \Phi^{(t)})$ is computed using Eqn.~\ref{eq:lD2}.
 		\STATE {\bf Step 2}:  For each $i$,
 		start from the current $Z_i $, run $l_\G$ steps of Langevin  dynamics to update $Z_i$, each of which follows Eqn.~\ref{eq:LangevinG}.
 		\STATE {\bf Step 3}: Update $\Theta^{(t+1)} = \Theta^{(t)} + \gamma_t \frac{\partial}{\partial \Theta}(L(\Theta^{(t)})-H(\Theta^{(t)}, \Phi^{(t+1)})) $,  where $\frac{\partial}{\partial \Theta}L(\Theta^{(t)})$ and $\frac{\partial}{\partial \Theta}H(\Theta^{(t)}, \Phi^{(t+1)})$ are computed using Eqn.~\ref{eq:weight} and \ref{eq:crossen} respectively.
 		\STATE Let $t \leftarrow t+1$
 		\UNTIL $t = T$
 	\end{algorithmic}
 \end{algorithm}

\section{Experiments}
In this section, we present the qualitative and quantitative results of the proposed method tested on four datasets widely used in image synthesis. The proposed method consistently obtains better quantitative performance with interpretable hierarchical representations learned. The code and results can be found at the project page~\footnote{{https://andyxingxl.github.io/Deep-Sparse-Generator/}}.

\textbf{Datasets:} We use the CelebA dataset~\cite{liu2015faceattributes}, the human fashion dataset~\cite{liuLQWTcvpr16DeepFashion}, the Stanford car dataset~\cite{KrauseStarkDengFei-Fei_3DRR2013}, the LSUN bedroom dataset~\cite{yu2015lsun}.
We train our proposed AND-OR model on the first 10k CelebA images as processed by OpenFace \cite{amos2016openface}, 78,979 human fashion images as done in~\cite{liuLQWTcvpr16DeepFashion},  the first 16k Stanford car images, and the first 100k bedroom images, all cropped to $64 \times 64$ pixels.

\textbf{Settings and baselines:} Table~\ref{table:network} summarizes architectures of the generator network and the energy-based network used in our experiments. We compare our model with state-of-the-art image synthesis methods including VAE~\cite{kingma2013auto}, DCGAN~\cite{radford2015unsupervised}, WGAN~\cite{arjovsky2017wasserstein}, CoopNet~\cite{coopnets}, CEGAN~\cite{dai2017calibrating}), ALI~\cite{dumoulin2016adversarially}, and ALICE~\cite{li2017alice}. We use the  Fr\'echet Inception distance (FID)~\cite{heusel2017gans} for evaluating the quality of generated images.  VAE (based on variational inference) and GAN (based on adversarial training) are two kinds of representatively generative models for image synthesis. CoopNet and CEGAN are both the energy-based models. However, the above traditional GAN and energy-based methods can only generate images and do not have the reconstruction power. Our model has both the generative and reconstruction power, so we further compare with ALI and ALICE, which are also GAN based models but also have the reconstruction power. The number of generated samples  for computing FID is the same as that of training set. We also compare the image reconstruction quality in terms per pixel mean square errors (MSE).


\begin{table}
	\centering
	\caption{Network architectures used in experiments. Upsample uses nearest neighbor interpolation. Downsample uses average pooling. LReLU is the leaky-ReLU with negative slope being 0.2. All convolution layers use kernels of size $3\times 3$ with the number of output channles listed in $(\cdot)$. The sparsity-inducing hyper-parameter $k$ is also given. }
	\vspace{2mm}
	\label{table:network}
	\scriptsize
	\begin{tabular}{c|c||c }
		\hline
		Layer  & Generator Network & Energy Based Network \\ \hline
		1 &  $Z\sim \mathcal{N}(0, I_{100})$ & Y,  ($64\times 64\times 3$) \\ \hline
		2 & FC, $(4\times 4\times 3200)$; $k=8$ & Conv+LReLU, $(64)$\\ \hline
		\multirow{3}{*}{3} & Upsample, 2  &  Downsample, 2 \\
		 & Conv+ReLU, $(512)$  & Conv+LReLU, $(64)$ \\
		 & Conv+ReLU, $(512)$; $k=\frac{8\times 8}{4}$ & Conv+LReLU, $(64)$ \\ \hline
		 \multirow{3}{*}{4} & Upsample, 2  & Downsample, 2  \\
		 & Conv+ReLU, $(256)$  &  Conv+LReLU, $(128)$\\
		 & Conv+ReLU, $(256)$; $k=\frac{16\times 16}{4}$ & Conv+LReLU, $(128)$ \\ \hline
		 \multirow{3}{*}{5} & Upsample, 2  & Downsample, 2 \\
		 & Conv+ReLU, $(128)$  & Conv+LReLU, $(256)$\\
		 & Conv+ReLU, $(128)$; $k=\frac{32\times 32}{4}$ & Conv+LReLU, $(256)$ \\ \hline
		 \multirow{3}{*}{6} & Upsample, 2  &  Downsample, 2 \\
		 & Conv+ReLU, $(64)$  & Conv+LReLU, $(512)$ \\
		 & Conv+ReLU, $(64)$ & Conv+LReLU, $(512)$\\ \hline
		 7 & Conv+Tanh, $(3)$ &  FC, $(1)$ \\ \hline
	\end{tabular}\vspace{-3mm}
\end{table}

 \begin{figure}[t!]
	\begin{center}	
		\includegraphics[width=\linewidth]{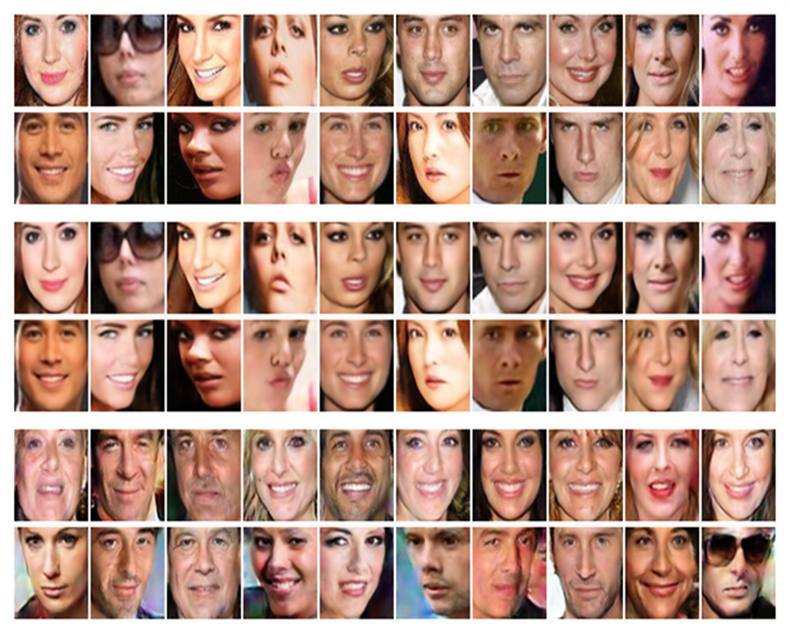}	
	\end{center} \vspace{-2mm}
		\caption{Results of image synthesis and reconstruction on CelebA. The first two rows show the original face images, the middle two rows show the reconstruction results, and the last two rows show the generated face images. The learned AND-OR tree model is illustrated in Figure~\ref{fig:aot_face}. } \label{fig:face} \vspace{-3mm}
\end{figure}

\begin{figure}[th!]
	\begin{center}	
		\includegraphics[width=\linewidth]{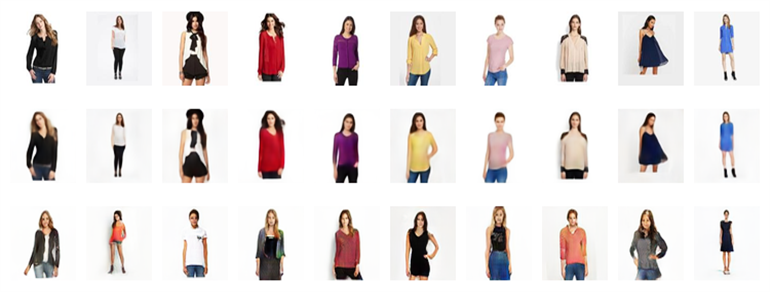}	\\
		\includegraphics[width=\linewidth]{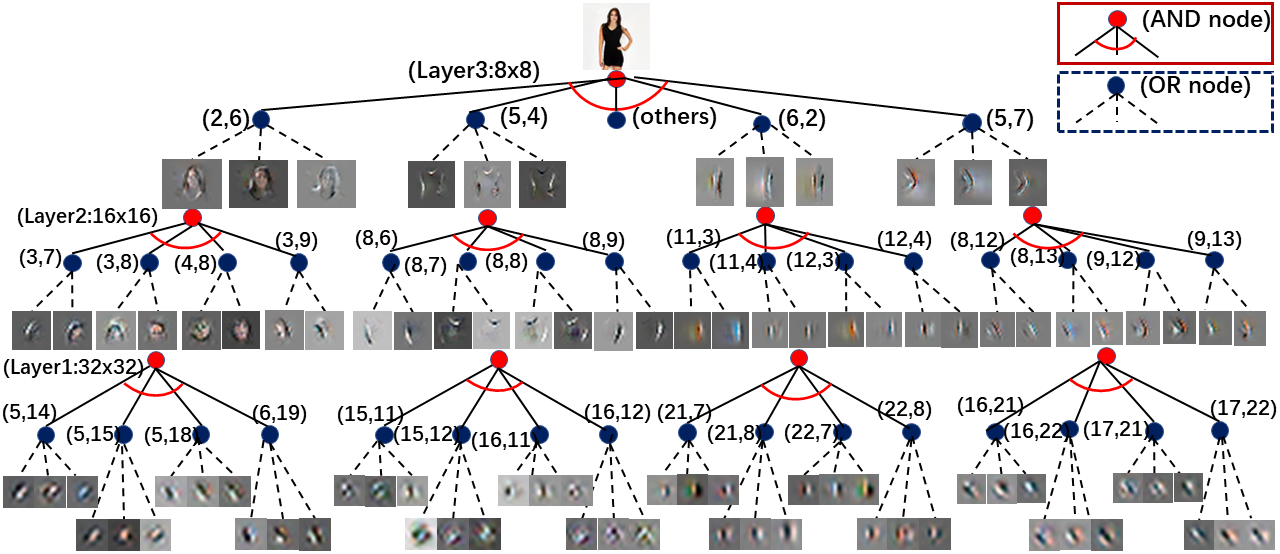}	
	\end{center}
	\caption{Results on the human fashion dataset. Top: the three rows show original images, reconstructed images and generated images respectively. Bottom: the learned AND-OR tree model.
	}
	\label{fig:human}\vspace{-3mm}
\end{figure}

\begin{figure}[th!]
	\begin{center}	
		\includegraphics[width=\linewidth]{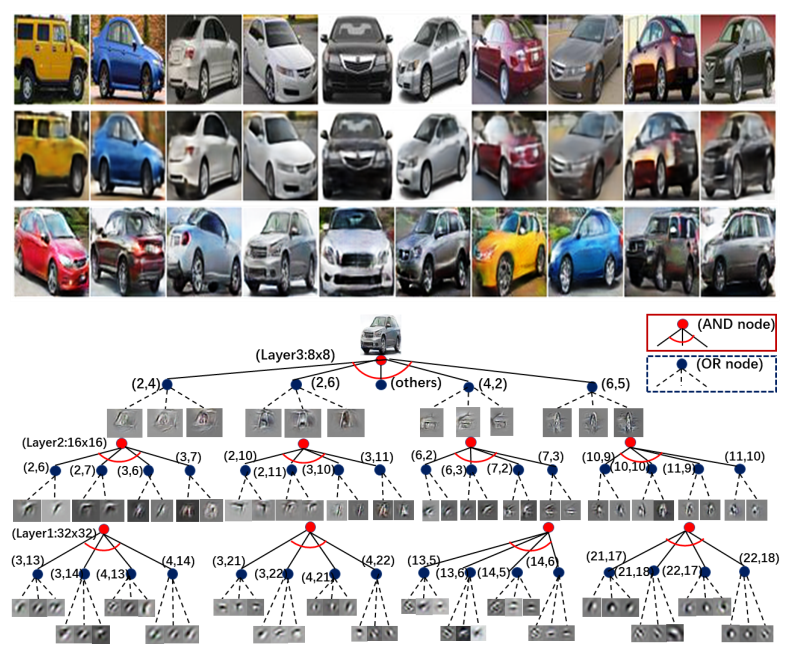}	
	\end{center}
	\caption{Results on the Stanford car dataset. Top: the three rows show original images, reconstructed images and generated images respectively. Bottom: the learned AND-OR tree model.
	}
	\label{fig:car}\vspace{-4mm}
\end{figure}

\begin{figure}[th!]
	\begin{center}	
		\includegraphics[width=\linewidth]{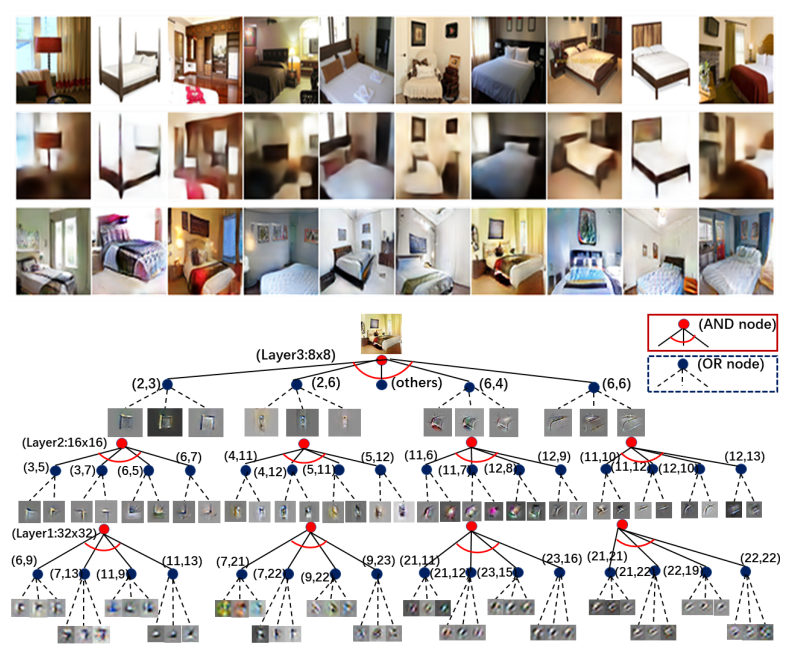}	
	\end{center}
	\caption{Results on the LSUN bedroom dataset. Top: the three rows show original images, reconstructed images and generated images respectively. Bottom: the learned AND-OR tree model.
	}
	\label{fig:bedroom} \vspace{-3mm}
\end{figure}

\begin{table*}[t!]
\centering
		\caption{Comparisons of the Fr\'echet inception distance (FID). Smaller FID is better. The last column $\Delta$ shows the improvement of our method over the runner-up method, WGAN.} \vspace{2mm}
		\label{table:FID}
 		\newsavebox{\tablebox}
 		\begin{lrbox}{\tablebox}
			\begin{tabular}{|l|c|c|c|c|c|c|c|c|c|}
				\cline{1-10}
				Datasets$\backslash$ Methods &VAE \cite{kingma2013auto}& DCGAN  \cite{radford2015unsupervised} & WGAN \cite{arjovsky2017wasserstein}  & CoopNet \cite{coopnets}& CEGAN  \cite{dai2017calibrating}&ALI  \cite{dumoulin2016adversarially}&ALICE \cite{li2017alice} &Ours & $\Delta$\\
				\cline{1-10}
				CelebA & 53.38  & 19.28 & \textit{18.85} & 28.49& 20.62& 30.53& 23.17&\textbf{16.62} & 2.32\\ \cline{1-10}
				HumanFashion & 27.94 & 10.82 &\textit{10.19} &15.39 & 11.14 &16.75&  12.56&  \textbf{8.65} & 1.44 \\
				\cline{1-10}
				Standford cars& 87.64  & 33.58 & \textit{31.62} &45.34 & 36.12&50.48 & 37.35& \textbf{28.36} & 2.26 \\
				\cline{1-10}
                LSUN bedrooms& 105.76  & 36.26 & \textit{33.81} &49.73 & 41.64&52.79 & 39.08& \textbf{29.70} & 4.11  \\
                \cline{1-10}
			\end{tabular}
 		\end{lrbox}
 		\scalebox{0.85}{\usebox{\tablebox}}
        \vspace{-4mm}
	\end{table*}

\subsection{Qualitative results}
Our AND-OR model is capable of joint image synthesis and reconstruction. Figure~\ref{fig:face} shows examples of reconstructed and generated face images. The top of Figure~\ref{fig:human},  Figure~\ref{fig:car} and Figure~\ref{fig:bedroom} show examples for human fashion images, car images and bedroom images respectively (where the learned AND-OR tree models are shown in the same way as Figure~\ref{fig:aot_face}). Both the reconstructed images and the generated images look sharp. The reconstructed images of bedroom (Figure~\ref{fig:bedroom}) look relatively blurrier. Bedroom images usually have larger variations which may entail more complicated generator and energy-based network architectures. We use the same architectures for all the tasks.

The learned AND-OR trees on the four datasets unfold the internal generation process with semantically meaningful internal basis functions learned (emerged). To our knowledge, this is the first work in image synthesis that learn interpretable image generation from scratch.  As we demonstrated in the Fig. ~\ref{fig:aot_face},~\ref{fig:human},~\ref{fig:car}, and ~\ref{fig:bedroom}, our model can mine semantically meaningful AND-OR Tree from the datasets automatically, which facilitate the generative process to be transparent and explainable. (E.g.) from Fig. ~\ref{fig:aot_face}, we have learned semantically meaningful parts, such as eyes, nose and mouth of a face. It is semantically meaningful, since, at the eye's location, the learned basis functions consist of different kinds of eyes, in other words, there are not other kinds of basis functions, such as nose and mouth, appearing at the eye's location. Moreover, we can find that the basis functions (subparts) of the eye's part are different from the basis functions from the nose and mouth parts. From Fig.~\ref{fig:human}, we can learn semantically meaningful parts, such as head, body, left arm and right arm of a human. For the car example in Figure~\ref{fig:car}, we can learn semantically meaningful parts, such as the car window, central pillar at the left side of the car, headlight and car tires. More interestingly, we observe that the primitive layers in different AND-OR trees share many common patterns similar to the Gabor wavelets and blob-like structures, which is also consistent with results in traditional sparse coding.

\begin{figure}[t!]
	\begin{center}	
		\includegraphics[width=\linewidth]{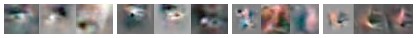}	\\
        \includegraphics[width=\linewidth]{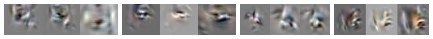}
	\end{center}
	\caption{Part-level basis functions learned from the traditional generator network on the face dataset. Comparing with the part-level (layer-3) basis functions learned with our sparsely connected AND-OR model as shown in Figure~\ref{fig:aot_face}, the traditional generator network cannot obtain meaningful internal basis functions. The first row shows the basis functions from the traditional generator network without sparsity-inducing constraint. The second row shows the results of the generator network combining with an energy network but without sparsity-inducing constraint.}
	\label{fig:densebasis} \vspace{-3mm}
\end{figure}

It is worth noting that the without the proposed sparsely connected AND-OR model, the traditional generator network cannot obtain these meaningful internal basis functions, as we can observe from Figure~\ref{fig:densebasis}. The reason is they utilize the distributed representations, the representative power for a single activation is weak. In contrast, in our sparsely connected AND-OR model, the energies are forced to be collected into a few activations to make the corresponding basis functions meaningful.

\subsection{Quantitative results}
The FID comparsions are summarized in Table~\ref{table:FID}.
The proposed method consistently outperforms the seven state-of-the-art image synthesis methods in comparisons. On the human fashion dataset, the images are nice and clean, our method obtains the least improvement by $1.44$. On the bedroom dataset, the images are much more complex with large structural and appearance variations, our method obtains the biggest improvement by $4.11$. We note that all the improvement are obtained with more interpretable representations learned in the form of AND-OR trees. This is especially interesting since it shows that jointly improving model performance and interpretability is possible.

We utilize per-pixel mean square error (MSE) to evaluate image reconstruction. Table~\ref{table:MSE} shows the comparisons with three state-of-the-art methods that are also capable of joint image synthesis and reconstruction (VAE \cite{kingma2013auto}, ALI \cite{dumoulin2016adversarially}, and ALICE \cite{li2017alice}). We do not compare with the variants of GANs and CoopNets since they usually can not perform joint image reconstruction. From the experimental results, the proposed method can not only obtain interpretable generative process, such as semantically meaningful object parts and primitives, but also obtain competitive or even better performance than the 7 famous traditional generative models on the synthesis and reconstruction tasks.

\subsection{Ablation studies}
In addition to the AND-OR tree visualization, we propose a simple method to evaluate the intepretability of learned basis functions (e.g., those at Layer 3, see Figure~\ref{fig:aot_face}).
We perform Template Matching between the learned basis functions with  training images using the fast normalized cross-correlation algorithm ~\cite{yoo2009fast}. Consider Layer 3 (a.k.a. object part level), if the learned basis functions contain meaningful local parts of the object, the matching score shall be high. We compare the Layer-3 basis functions learned with and without the proposed sparsity-inducing approach respectively (i.e., Eqn.~\ref{eq:factor_sparse} vs Eqn.~\ref{eq:factor}). The results of the mean matching scores are summarized in Table~\ref{table:matchscore}. The proposed method significantly outperforms the counterpart. The results verify that the proposed method can learn meaningful basis functions for better model interpretability.

\begin{table}[t!]
	\centering
	\caption{Comparisons of the per-pixel mean square error (MSE). Smaller MSE is better.}
	\vspace{3mm}
	\label{table:MSE}
	\begin{lrbox}{\tablebox}
		\begin{tabular}{|l|c|c|c|c|}
			\cline{1-5}
			Datasets$\backslash$ Methods &VAE \cite{kingma2013auto} &  ALI  \cite{dumoulin2016adversarially}&ALICE \cite{li2017alice}&Ours\\
			\cline{1-5}
			CelebA & 0.016  & 0.132 & 0.019 & \textbf{0.011} \\
			\cline{1-5}
			HumanFashion & 0.033 & 0.28  & 0.043 & \textbf{0.024} \\
			\cline{1-5}
			Standford cars& 0.081  & 0.563 & 0.078 &\textbf{0.054}  \\
			\cline{1-5}
			LSUN bedrooms& 0.154  & 0.988 & 0.127 &\textbf{0.097}  \\
			\cline{1-5}
		\end{tabular}
	\end{lrbox}
	\scalebox{0.85}{\usebox{\tablebox}} \vspace{-2mm}
\end{table}

\begin{table}[t!]
\centering
		\caption{Evaluation of interpretability. Comparisons of the matching scores using the fast normalized cross-correlation algorithm between the generator without sparsity and the proposed sparse activated generator.}
		\vspace{3mm}
		\label{table:matchscore}
 		\begin{lrbox}{\tablebox}
			\begin{tabular}{|l|c|c|c|c|}
				\cline{1-5}
				Methods$\backslash$Datasets  & CelebA & HumanFashion &  Cars   &  Bedroom\\
				\cline{1-5}
				w/o sparsity & 0.33  & 0.29 & 0.31 & 0.23  \\
				\cline{1-5}
				w/ sparsity & \textbf{0.83} & \textbf{0.81} & \textbf{0.76} & \textbf{0.72} \\
				\cline{1-5}
			\end{tabular}
 		\end{lrbox}
 		\scalebox{0.83}{\usebox{\tablebox}}
	 \label{table:matchscore} \vspace{-2mm}
	\end{table}
	
\section{Conclusion}
    This paper proposes interpretable image synthesis by sparsifying generator network to induce a hierarchical compositional AND-OR model.  The proposed method is built on the vanilla generator network, which inherits the implicitly hierarchal grammar of the convolutional network. The AND-OR model of sparsely connected nodes emerges from the original densely connected generator network when sparsity-inducing functions are introduced. Our work unifies the top-down generator network and sparse coding model, can learn interpretable dictionaries at multiple layers, moreover, learn the compositional structures and primitives automatically from data. In training, we further recruit an energy-based model and we jointly train the generator model and the energy-based model as actor and critic. The resulting AND-OR model is capable of image synthesis and reconstruction. The results show that meaningful and interpretable hierarchical representations are learned with better qualities of image synthesis and reconstruction obtained than baseline methods.

 \section*{Acknowledgments}
    The work of X. Xing, S.-C. Zhu and Y. Wu is supported by DARPA XAI project N66001-17-2-4029; ARO project W911NF1810296; and ONR MURI project N00014-16-1-2007; and Extreme Science and Engineering Discovery Environment (XSEDE) grant ASC170063. The work of X. Xing is also supported by Natural Science Foundation of China No. 61703119, Natural Science Fund of Heilongjiang Province of China No. QC2017070, and Fundamental Research Funds for the Central Universities No. 3072019CFT0402. The work of T. Wu is supported by NSF IIS-1909644 and ARO grant W911NF1810295.

{\small
\bibliographystyle{ieee_fullname}
\bibliography{cvpr2020version}
}

\end{document}